\documentclass{article}

\usepackage{arxiv}
\usepackage{amsfonts,booktabs,multirow,lineno,hyperref,xcolor,amsmath}
\usepackage{blkarray, bigstrut}
\usepackage{booktabs}
\newcommand\topstrut[1][1.2ex]{\setlength\bigstrutjot{#1}{\bigstrut[t]}}
\newcommand\botstrut[1][0.9ex]{\setlength\bigstrutjot{#1}{\bigstrut[b]}}

\usepackage[utf8]{inputenc} 
\usepackage[T1]{fontenc}    
\usepackage{hyperref}       
\usepackage{url}            
\usepackage{booktabs}       
\usepackage{amsfonts}       
\usepackage{nicefrac}       
\usepackage{microtype}      
\usepackage{lipsum}
\usepackage{graphicx}
\graphicspath{ {./images/} }

\usepackage{amsmath,amssymb}
\usepackage{adjustbox}
\usepackage{caption}
\usepackage{subfigure}
\usepackage{multirow}
\usepackage[english]{babel}
\usepackage{amsthm}
\newtheorem{definition}{Definition}

\newtheorem{proposition}{Proposition}

\newcommand*{\QEDB}{\null\nobreak\hfill\ensuremath{\square}}

\title{The foundations of cost-sensitive causal classification}

\author{
Wouter~Verbeke\\
Faculty of Economics and Business \\
Katholieke Universiteit Leuven\\
Leuven, Belgium\\
\texttt{wouter.verbeke@kuleuven.be}
\And
Diego~Olaya \\
Data Analytics Laboratory \\
Solvay Business School\\
Vrije Universiteit Brussel\\
Brussels, Belgium\\
\texttt{diego.olaya@vub.be}
\And
Jeroen~Berrevoets \\
Department of Applied Mathematics \\
and Theoretical Physics \\
University of Cambridge \\
Cambridge, United Kingdom \\
\texttt{jeroen.berrevoets@vub.be}
\And
Sam~Verboven \\
Data Analytics Laboratory \\
Solvay Business School\\
Vrije Universiteit Brussel\\
Brussels, Belgium\\
\texttt{sam.verboven@vub.be}
\And
Sebasti\'an Maldonado \\
School of Economics and Business \\
University of Chile\\
Santiago, Chile\\
\texttt{sebastianm@fen.uchile.cl}
}

\begin{document}
\maketitle

\begin{abstract}
Classification is a well-studied machine learning task that concerns the assignment of instances to a set of outcomes. Classification models support the optimization of managerial decision-making across a variety of operational business processes. 
For instance, customer churn prediction models are adopted to increase the efficiency of retention campaigns by optimizing the selection of customers intended to be targeted. 
Cost-sensitive as well as causal classification methods have independently been proposed to improve the performance of classification models. The former approach considers the benefits and costs of correct and incorrect classifications, such as the benefit of a retained customer, whereas the latter estimates the causal effect of a treatment, such as a retention campaign, on the outcome of interest. 
This study integrates cost-sensitive and causal classification by developing a unifying evaluation framework. The framework encompasses a range of existing and novel performance measures for evaluating both causal and conventional classification models in a cost-sensitive as well as cost-insensitive manner. Causal classification performance measures as defined in terms of the proposed framework are shown to instantiate to well-known conventional classification performance measures. 
The framework instantiates to application-specific cost-sensitive performance measures that have recently been proposed for evaluating customer retention and response uplift models, and allows maximizing profitability when adopting a causal classification model for decision-making. The proposed framework paves the way toward the development of cost-sensitive causal learning methods.
\end{abstract}


\section{Introduction}

A range of business decision-making processes across management domains are supported by classification models that are learned from historical data through machine learning methods. Classification is a well-studied machine learning task that concerns the assignment of instances to a predefined set of outcome classes. For instance, a customer churn prediction model classifies customers as future churners or non-churners, which allows customers who are likely to churn to be selected for targeting in a retention campaign \cite{coussement2010improved,keramati2014improved}.

Two improvements to standard classification methods have been proposed to improve the performance of classification models:

\begin{enumerate}
    \item \textbf{Cost-sensitive classification methods take into account asymmetric costs and benefits related to incorrect and correct classifications, respectively} \cite{elkan2001foundations}. For instance, a future churner who is correctly identified by a churn prediction model may be retained by the retention campaign and yield a benefit. Incorrectly classifying a churner as a non-churner, on the other hand, will involve a cost. This cost, moreover, is larger than the cost of incorrectly classifying a non-churner as a churner as it is usually more costly to lose a customer than to target a non-churner in a retention campaign. Cost-sensitive classification methods that take into account these imbalanced costs and benefits better align with the true business objective, i.e., maximizing profit, as shown in the literature \cite{verbraken2012novel,petrides2020cost}.
    \item \textbf{Causal classification methods predict the net effect of a treatment on an outcome of interest} \cite{wager2018}. For instance, uplift models allow estimation of the change in the probability of a customer to churn when targeted with a retention campaign, which aligns with the business objective of maximizing the reduction in churn rather than maximizing the number of correctly identified churners. It has been shown in the literature that causal learning for maximizing the effect of marketing campaigns significantly increases profitability \cite{ascarza2018retention, devriendt2019,gubela2019response}. 
\end{enumerate}

Whereas cost-sensitive classification methods merely extend upon standard, i.e., cost-insensitive, classification methods, causal classification methods involve a paradigm shift. Essentially, when causal learning is applied to support decision-making, the aim is to learn a simulation model from the data. Hence, causal learning can be considered a type of prescriptive analytics and part of the realm of operations research, since the objective is to optimize decision-making. 

\textbf{Research gap.} Both types of methods have been shown in the literature to substantially improve the use of classification models for supporting business decision-making. \cite{hoppner2018profit,ascarza2018retention}. To the best of our knowledge, however, no methods have been proposed in the literature that combine cost-sensitive and causal learning. 
    
\textbf{Contributions.} Therefore, in this article, we introduce and define cost-sensitive causal learning as a specialized machine learning task and develop a framework that integrates cost-sensitive and cost-insensitive, as well as causal and conventional classification. 
To this end, we formalize the foundations of cost-sensitive evaluation of conventional classification models by introducing the effect matrix as the difference between the confusion matrix of a classification model and a baseline model of choice. This approach results in an invariant and hence unambiguous definition of the cost-benefit matrix.

Subsequently, we extend upon these foundations and develop a framework for evaluating causal classification models, by introducing the causal counterparts of the confusion matrix, cost-benefit matrix and newly proposed effect matrix. This framework allows formalizing the scattered body of literature on cost-insensitive performance measures as well as defining a range of novel performance measures, both cost-sensitive and cost-insensitive, for evaluating causal classification models. The framework is shown to instantiate to a range of existing performance measures for evaluating conventional classification models, hence embodying a general theory encompassing the evaluation of both conventional and causal, cost-insensitive and cost-sensitive classification. 

The practical use of this approach in business decision-making is illustrated by instantiating the framework to two application-specific profit measures that have been recently proposed in the literature for evaluating customer churn \cite{devriendt2019} and customer response \cite{gubela2019response} uplift models. 

\textbf{Outline.} In Section \ref{sec:costsensclas}, a mathematical framework for evaluating classification models in a cost-insensitive and a cost-sensitive manner is introduced. This section presents a formalized approach to relative performance evaluation, i.e., when compared to a baseline model of choice. 
Next, in Section \ref{sec:causalclasperf}, we develop a similar framework for evaluating the performance of causal classification models, which incorporates a range of existing performance measures and allows introducing a novel measure. Then, in Section \ref{sec:costsenscausal}, the framework is extended to facilitate a cost-sensitive evaluation of causal classification models. Section \ref{sec:instantiation} 
illustrates the use of the framework for business decision-making and Section \ref{sec:conclusions} presents conclusions and future research opportunities.

\section{Cost-sensitive classification performance}\label{sec:costsensclas}

\subsection{Fundamentals of cost-sensitive classification performance}\label{sec:fund}

Classification is a common machine learning task that concerns the assignment of instances $x$ from the instance space $X \subseteq \mathbb{R}^n$ to a class or outcome $Y$. In this article, we focus on binary classification, i.e., $Y \in \{0,1\}.$ By convention, we refer to $Y=1$ as the positive class or outcome and to $Y=0$ as the negative class outcome. Instances with outcome $Y=1$ are called positive instances, and those with $Y=0$ are called negative instances. A classification model is defined as a function $m: X \rightarrow [0,1]$ that maps instances $x$ to a positive outcome probability, $P(Y=1|x)$ or $p_1$ in short, and a negative outcome probability, $P(Y=0|x) = 1-P(Y=1|x)$, or $P_0$ in short. A predicted class, $\hat{Y} \in \{0,1\}$, is obtained by setting a classification threshold $\phi$. Instances with a positive outcome probability below the threshold, $p_1<\phi$, are classified in the negative class, and instances with a score above the threshold are classified in the positive class. 

Classification methods learn a classification model from a data set $\mathcal{D} = \{(x_i, y_i)\ : i=1,...N\}$, where $N=\lvert \mathcal{D} \rvert$ is the number of instances in the data set. $\mathcal{D}_k$ is the subset of instances in class $k\in Y$, where $N_k = \lvert \mathcal{D}_k \rvert$ is the number of instances in class $k$ in data set $\mathcal{D}$, and $\pi_k=N_k/N$ is the prior probability of instances belonging to class $k$. Since $\sum_k N_k=N$, we have $\sum_k \pi_k=1$, which for the binary case yields $\pi_0+\pi_1=1$. We call $\pi_0$ the negative class proportion and $\pi_1$ the positive class proportion. 
The positive outcome probability density for class $k$ will be denoted by $f_k(p_1)$ and the cumulative distribution by $F_k(p_1)$. Hence, $F_1(\phi)=\int_{-\infty}
^\phi f_1(p_1)dp_1 = P(p_1\leq \phi \lvert y = 1)$ is the proportion of positive instances with a positive outcome probability $p_1$ below the threshold $\phi$.

The confusion matrix, $\mathbf{CF}$, summarizes the number of correctly and incorrectly classified instances of the negative and positive class for a threshold $\phi$: 
\begin{equation} \label{eq:CF}
\mathbf{CF} := \hspace{0.5cm}
    \begin{blockarray}{cccc}
        \BAmulticolumn{2}{c}{\text{Predicted class}} & & \\
        \hat{Y}=0 & \hat{Y}=1 & & \\
        \addlinespace
        \begin{block}{[cc]cc}
            \topstrut \pi_0F_0(\phi) & \pi_0\big(1 - F_0(\phi)\big) & \multirow{2}{*}{$Y=0$} & \multirow{4}{*}{Outcome class}\\
            \text{\small{(True negatives)}} & \text{\small{(False positives)}} & & \\
            \addlinespace
            \pi_1F_1(\phi) & \pi_1\big(1 - F_1(\phi)\big) & \multirow{2}{*}{$Y=1$} &\\
            \botstrut \text{\small{(False negatives)}} & \text{\small{(True positives)}} & & \\
      \end{block}
    \end{blockarray}
\end{equation}
The confusion matrix as defined in Equation \eqref{eq:CF} reports the number of correctly and incorrectly classified instances for each class as a proportion of the total number of instances $N$ and as a function of the class proportion $\pi_k$ and the cumulative class distributions $F_k(p_1)$. For instance, the lower left element, $\mathbf{CF}_{10}$, is the proportion of positive instances that are incorrectly classified in the negative class for threshold $\phi$, which are called false negatives. Various performance measures for evaluating classification models can be defined in terms of the confusion matrix, e.g.,
\begin{equation}
    \text{Accuracy} := \pi_0 F_0(\phi) + \pi_1 \big(1-F_1(\phi)\big), 
\end{equation}
\begin{equation}\label{eq:sens}
    \text{Sensitivity} := 1 - F_1(\phi), 
\end{equation}
\begin{equation}
    \text{Specificity} := F_0(\phi). 
\end{equation}

The receiver operating characteristic (ROC) curve plots the sensitivity, $1-F_1(\phi)$, for all possible thresholds on the Y-axis versus the false discovery rate, $1-F_0(\phi)$, on the X-axis. As a measure for assessing the performance of a classification model, the area under the ROC curve (AUC) is often used as a performance measure \cite{hand2009measuring}:
\begin{equation}\label{eq:AUC}
    \text{AUC} := \int_{0}^{1} \big(1-F_1(p_1)\big)f_0(p_1)\text{d}p_1.
\end{equation}
The gains curve, which is also denominated the cumulative accuracy profile (CAP) or Lorenz curve, is an alternative to the ROC curve and plots the sensitivity as a function of the positive prediction rate, $\eta$, which is defined as the proportion of instances that are classified in the positive class:
\begin{equation}
    \eta =  \pi_0 \big(1-F_0(\phi)\big) + \pi_1 \big(1-F_1(\phi)\big).
\end{equation}

The positive prediction rate is a function of the classification threshold. The inverse function yields the classification threshold that is equivalent with a positive prediction rate,  i.e.,  $\eta^{-1}=\phi$.

The Gini coefficient of a classification model is equal to the ratio of the area in between the gains curves of the model and a random classification model, and the area in between the gains curves of the perfect model and a random model:
\begin{equation}\label{eq:Gini}
    \text{Gini} := \frac{2\int_{0}^{1} \big(1-F_1(\eta^{-1})\big)\text{d}\eta - 1}{1-\pi_1}.
\end{equation}

A third popular method for visualizing model performance is the lift curve, which plots the lift as a function of the positive prediction rate. The lift at a positive prediction rate, $\lambda(\phi)$, is defined as the ratio of the proportion of positive outcomes among the predicted positive instances and the overall proportion of positive instances, i.e., the prior positive class probability, $\pi_1$:
\begin{equation}\label{eq:Lift}
    \lambda(\phi):=\frac{\frac{\pi_1 \big(1-F_1(\phi)\big)}{\phi}}{\pi_1} = \frac{\big(1-F_1(\phi)\big)}{\phi} .
\end{equation}

In practical applications, the four types of outcomes in the confusion matrix may have very different monetary implications. False positive and false negative classifications will typically yield different costs resulting from incorrect decision-making, whereas true positive and true negative classifications usually involve a different benefit. The costs and benefits associated with the four segments in the confusion matrix can be specified in a cost-benefit matrix, $\textbf{CB}$:
\begin{equation} \label{eq:CB}
    \begin{blockarray}{ccccc}
        & \BAmulticolumn{2}{c}{\text{Predicted class}} & & \\
        & \hat{Y}=0 & \hat{Y}=1 & & \\
        \addlinespace
        \begin{block}{c[cc]cc}
            \topstrut \multirow{2}{*}{$\mathbf{CB}$ :=} & cb_{00} & cb_{01} & Y=0 & \multirow{2}{*}{Outcome class}\\
           \addlinespace
            & cb_{10} & cb_{11} & Y=1 & &\\
        \end{block}
    \end{blockarray}
\end{equation}

The cost or benefit of classifying an instance from class $i$ in class $j$ is denoted by $cb_{ij}$. By convention, a positive value in the cost-benefit matrix represents a benefit and a negative value represents a cost. In Section \ref{sec:relperf}, we propose a formalized definition of the cost-benefit matrix, which can be instantiated to various definitions that can be found in the literature.

By summing the products of the elements of the confusion matrix and the corresponding elements in the cost-benefit matrix, we obtain the classification profit per instance, which we denote by $P$:
\begin{equation}\label{eq:profClass}
    P := \Sigma_{i,j}(\mathbf{CF}_{ij} \circ \mathbf{CB}_{ij}),
\end{equation}
where $\Sigma_{i,j}(\mathbf{A_{ij}})$ is the sum of all elements of matrix $\mathbf{A}$ and $\circ$ is the Hadamard or elementwise product operator. 

As a measure for evaluating the performance of a classification model, \cite{verbeke2012new} proposed the use of the maximum profit measure, $MP$, which is defined as:
\begin{equation}\label{eq:MP}
    MP := \max_{\forall \phi} \big(P(\phi; \mathbf{CB})\big) = P(\phi^{*}; \mathbf{CB}),
\end{equation}
where $\phi^{*}$ is the profit-maximizing threshold. 
As an extension to the $MP$ measure, \cite{verbraken2012novel} proposed the expected maximum profit ($EMP$) measure, which acknowledges that the cost and benefit parameters may not be exactly known \cite{de2020cost} or may vary in time or across instances following some joint probability distribution, $h(\mathbf{CB})$. The $EMP$ measure is defined as:
\begin{equation}\label{eq:EMP}
    EMP := \int_{-\infty}^{\infty} P(\phi^{*};\mathbf{CB}) \cdot h(\mathbf{CB}) d\mathbf{CB}.
\end{equation}
The EMP measure achieves a robust evaluation by allowing the consideration of random cost and benefit parameters in evaluating a classification model, reflecting the uncertainty of the exact values of these parameters or, when benefits and costs are random, their distribution. 

\subsection{Relative cost-sensitive classification performance}\label{sec:relperf}

The performance of a classification model can be measured in an absolute sense, for instance, by means of $P$, $MP$ or $EMP$, as defined in Section \ref{sec:costsensclas}. The result is to be interpreted relative to a value of zero, which is obtained when no instances are classified, i.e., when no outcomes occur\footnote{$P$, $MP$ and $EMP$ may as well be equal to zero in specific cases, when total costs and benefits are perfectly in balance.}. In the literature, however, the performance of a classification model is often, implicitly, evaluated in a relative manner, i.e., in comparison to a meaningful baseline model. Three common baseline models encountered in the literature are (1) a perfect model, which classifies all instances correctly and is typically adopted in cost-sensitive learning \cite{elkan2001foundations}, (2) a dummy model, which classifies all instances in one class and is typically adopted in profit-driven learning \cite{hoppner2018profit}, and (3) a random model, which classifies all instances randomly and is typically adopted in cost-insensitive learning.

For calculating the profit relative to these baseline models, the cost-benefit matrix is typically adapted. 
For calculating the profit relative to a perfect model, a hollow cost-benefit matrix is adopted where $cb_{00} = cb_{11} = 0$, yielding a zero benefit for correctly classified instances. Hence, only the cost of errors is taken into account in calculating measures such as the expected loss \cite{orallo2012}, the total misclassification cost ($TMC = N \cdot P$) or the average misclassification cost or loss ($AMC = P)$ \cite{verbeke2017profit}.

For calculating the profit relative to a dummy model that classifies all instances in the positive class, a cost-benefit matrix is used with  the benefit of a true positive and the cost of a false positive equal to zero, $cb_{11} = cb_{01} = 0$. Hence, no benefit or cost is considered for instances that are classified in the positive class when comparing performance to a model that classifies all instances in the positive class.

No straightforward alteration to the cost-benefit matrix can be made that allows calculating the profit relative to a random model. To this end, an alternative approach is needed.

As an alternative approach to altering the cost-benefit matrix, which appears to be the common practice in the literature as well as in software implementations, we define the relative profit for calculating the profit relative to a baseline model of choice, $P_r$, as the difference between the absolute profit that is obtained by the classification model, $P$, and the absolute profit that is obtained by the baseline model of choice, $P_b$:
\begin{definition}
\begin{equation}\label{eq:baseline}
P_r := P - P_b 
\end{equation}
\end{definition}

If we specify the baseline model of choice in terms of its confusion matrix, $\mathbf{CF_b}$, then we can elaborate the above equation using Equation \eqref{eq:profClass} to calculate the absolute profit of both the classification model, $P$, and the baseline model, $P_b$, which yields the following formula for $P_r$:
\begin{align}\label{eq:baseline2}
P_r &= \sum_{i,j} (\mathbf{CF}_{ij} \circ \mathbf{CB}_{ij}) - \sum_{ij} (\mathbf{CF_b}_{,ij} \circ \mathbf{CB}_{ij}), \nonumber\\
&= \sum_{i,j} \big((\mathbf{CF}_{ij}-\mathbf{CF_b}_{,ij}) \circ \mathbf{CB}_{ij}\big), \nonumber\\
&= \sum_{i,j} (\mathbf{E}_{ij} \circ \mathbf{CB}_{ij}).
\end{align}

\begin{definition}
The \textbf{effect matrix} is equal to the difference between the confusion matrix of a classification model, $\mathbf{CF}$, and the confusion matrix of a baseline model $\mathbf{CF_b}$. 
\begin{equation}\label{eq:E}
    \mathbf{E} := \mathbf{CF}-\mathbf{CF_b}
\end{equation}
\end{definition}

\begin{proposition}\label{proposition:E}
\begin{equation*}
    \sum_{i,j}\mathbf{E}_{ij} = 0
\end{equation*}
\end{proposition}
The proof of Proposition \ref{proposition:E} is provided in the Appendix. Proposition \ref{proposition:E} embodies an intuitive quid pro quo condition; in comparison with the baseline, an increase in some types of outcomes (e.g., the number of true positives and true negatives) must be offset by a decrease in outcomes of the other types (e.g., the number of false positives and false negatives).

The baseline models that are frequently used in the literature for relative performance evaluation are the following. The baseline confusion matrix for a perfect model that classifies all instances correctly:
\begin{equation}
\mathbf{CF_b}^{\text{perf}}=
    \begin{bmatrix}
\pi_0 & 0\\
0 & \pi_1
    \end{bmatrix}.
\end{equation}
The baseline confusion matrix for a dummy model that classifies all instances in the positive class:
\begin{equation}
\mathbf{CF_b}^{\text{pos}}=
    \begin{bmatrix}
0 & \pi_0 \\
0 & \pi_1
    \end{bmatrix}.
\end{equation}
Similarly, for a dummy model that classifies all instances in the negative class:
\begin{equation}
\mathbf{CF_b}^{\text{neg}}=
    \begin{bmatrix}
\pi_0 & 0\\
\pi_1 & 0
    \end{bmatrix}.
\end{equation}
The baseline confusion matrix for a random model, which classifies instances in a class with a probability equal to the prior class probability:
\begin{equation}
\mathbf{CF_b}^{\text{rand}}=
    \begin{bmatrix}
\pi_0 \cdot \pi_0 & (1-\pi_0) \cdot \pi_0\\
(1-\pi_1) \cdot \pi_1 & \pi_1 \cdot \pi_1
    \end{bmatrix}.
\end{equation}
The baseline confusion matrix for obtaining the absolute profit, finally, is the zero matrix:
\begin{equation}
\mathbf{CF_b}^{\text{abs}}=
    \begin{bmatrix}
0 & 0\\
0 & 0
    \end{bmatrix},
\end{equation}
yielding $P_b = 0$ and $P_r = P$. 

Equation \eqref{eq:baseline} in our view is a conceptually sounder approach than adjusting the cost-benefit matrix depending on the baseline model that is adopted. Moreover, the proposed approach straightens the interpretation and facilitates the specification of the cost-benefit parameters, $cb_{ij}$. In adopting the approach of adjusting the cost-benefit matrix, users are often inclined to take into account opportunity costs. For instance, when a positive instance is misclassified, often the missed benefit for classifying a positive instance correctly is included in the cost of a false negative. This practice, however, is incorrect and leads to double-counting this benefit, as becomes apparent with the proposed approach. By requiring an explicit baseline model to be selected and avoiding adjustments to be made to the cost-benefit matrix, the proposed approach ensures a correct and interpretable result. We do acknowledge that a full and precise specification of the $\mathbf{CB}$ matrix (or the joint probability distribution $h(\mathbf{CB})$) in practice may be difficult to obtain\cite{orallo2012}, yet it is indispensable for an accurate evaluation and optimal decision-making.

\section{Causal classification performance}\label{sec:causalclasperf}

In this section, we introduce causal classification and extend the framework for evaluating the performance of conventional classification models toward causal classification models. 

\subsection{Causal classification}

\begin{definition}\label{def:cc}
\textbf{Causal classification} involves the assignment of instances $x$ from the instance space $X \subseteq \mathbb{R}^n$ to a class or outcome $Y$, conditional on the treatment, $W$, that is applied. 
\end{definition}
This \textit{broad} definition of causal classification aligns with the Neyman-Rubin potential outcome framework \cite{rubin1974estimating,splawa1990application}. It goes beyond the \textit{narrow} definition in \cite{fernandez2019causal} to facilitate the formulation of a general evaluation framework for cost-sensitive and cost-insensitive evaluation across conventional and causal classification.

In this article, we focus on double-binary causal classification, i.e., $Y=\{0,1\}$ and $W = \{0,1\}$. We refer to treatment $W=0$ as the negative treatment and to treatment $W=1$ as the positive treatment. Note that often the negative treatment represents a do-nothing scenario. The outcome of instance $x$ for treatment $w$ is denoted by $Y_w$. In this setup, double-binary causal classification implies the estimation of the causal effect of a binary treatment on a binary outcome of interest at the level of an individual instance, i.e., the individual treatment effect (ITE). The ITE is defined here as the difference in outcomes for the positive and the negative treatment, i.e., $ITE = Y_1 - Y_0$. Thus, given $Y=\{0,1\}$, we have that $ITE \in \{-1,0,1\}$. 

Causal classification has been addressed in the literature as uplift modeling \cite{radcliffe1999differential,kane2014mining}, heterogeneous or individual treatment effect estimation \cite{imai2013estimating,wager2018}, individualized treatment rule learning \cite{qian2011performance} and conditional average treatment effect estimation \cite{shalit2017estimating}. Causal classification is an instance of counterfactual estimation and causal learning \cite{pearl2009causality}. A subtle difference across various approaches is in the objective, which can either be to predict or to explain \cite{shmueli2010explain}, i.e., to optimize the treatments that are applied at the individual instance level, which is the objective in this article, or to establish and measure the strength of causal relations among a set of variables, which is the objective in causal inference and structural equation modeling \cite{fernandez2019causal}. 

From a technical perspective, causal classification differs from conventional classification since the outcome variable, i.e., the ITE, cannot be observed. In real-world applications, only one treatment at a time can be applied to a subject. Hence, we cannot observe the difference in outcome, i.e., the individual treatment effect, for applying the different treatments. This problem is called the fundamental problem of causal inference \cite{holland1986statistics} and has given rise to a specialized class of methods for learning causal classification models, which estimate the ITE under the assumptions of ignorability, common support and stable unit treatment value (SUTVA) \cite{rubin1978bayesian,rosenbaum1983central}. 

For learning a causal classification model with a binary treatment variable, in principle, a causal classification method requires two random samples that are obtained by means of a randomized controlled trial and that are independent and identically distributed and representative for the population of interest\footnote{Various methods for learning causal models from observational data that suffer from selection bias, however, have been proposed in the literature, for instance, by sampling or learning a balanced representation using propensity score matching \cite{caliendo2008some} or domain adversarial training \cite{ganin2016domain}}. A first sample concerns instances that received the positive treatment, $W=1$. This sample is called the treatment sample and will be denoted by $\mathcal{D}^T$. A second sample concerns instances that received the negative treatment, $W=0$ and is called the control sample, which is denoted by $\mathcal{D}^C$. Then $\mathcal{D}=\{\mathcal{D}^T,\mathcal{D}^C\}=\{(x_i,w_i,y_i):i=1,..., N \}$ with $N = \lvert \mathcal{D} \rvert$ the number of instances in the data set, $N^T= \lvert \mathcal{D}^T \rvert$ and  $N^C= \lvert \mathcal{D}^C \rvert$ the number of instances in the treatment and control sample, respectively. The number of instances of class k in the treatment and control sample is denoted by $N^T_k$ and $N^C_k$, respectively. The prior class probabilities are denoted by $\pi_k^T=N^T_k/N^T$ for the treatment sample, and with $\pi_k^C=N^C_k/N^C$ for the control sample; by definition, it holds that $\pi_1^T + \pi_0^T = 1$ and $\pi_1^C + \pi_0^C = 1$. 

A causal classification model is a function $\dot{m}: X \to t \in [-1,1]$, where $t$ is the estimated ITE:
\begin{equation} \label{eq:estimated_ITE}
    t := P(Y=1|x,W=1) - P(Y=1|x,W=0).
\end{equation}

The estimated ITE allows ranking instances from the largest estimated positive individual treatment effect to the largest negative individual treatment effect. The challenge in evaluating causal classification models is to assess the accuracy of this ranking with respect to the effect of the treatment on the outcomes of the instances. To causally classify instances and to decide on the treatment that is to be applied for individual subjects depending on the estimated ITE, a causal classification threshold $\dot{\phi}$ is to be set. A causal classification model with a threshold is called a causal classifier. Instances with estimated ITE below the threshold, $t<\dot{\phi}$, are classified in the negative treatment class; i.e., treatment $W=0$ is prescribed. Instances with estimated ITE above the threshold, $t>\dot{\phi}$, are classified in the positive treatment class; i.e., treatment $W=1$ is prescribed. 

The estimated ITE density functions of negative and positive class instances in the treatment sample are denoted by $f_0^T(t)$ and $f_1^T(t)$, respectively, whereas the cumulative distributions are denoted by $F_0^T(t)$ and $F_1^T(t)$. For the control sample, we have $f_0^C(t)$, $f_1^C(t)$, $F_0^C(t)$ and $F_1^C(t)$. The proportion of instances in a sample with a positive and negative outcome that have been classified in the negative and positive treatment class is reported in the sample causal confusion matrix.

\begin{definition}
The \textbf{sample causal confusion matrix}, $\dot{\mathbf{CF}}^S$, reports the proportion of instances with a positive and a negative outcome that have been classified in the negative and positive treatment class by a causal classification model with threshold $\dot{\phi}$ for sample $S$. 
\end{definition}

\begin{proposition}\label{propositionPO}
$\Sigma_{i,j}(\dot{\mathbf{CF}}^S)=1$
\end{proposition}

\noindent The proof of proposition \ref{propositionPO} is provided in the Appendix.

\noindent The treatment sample causal confusion matrix is denoted by $\dot{\mathbf{CF}}^T$ and defined as:
\begin{equation} \label{eq:POT}
\dot{\mathbf{CF}}^T := \hspace{0.5cm}
    \begin{blockarray}{ccccc}
        \BAmulticolumn{2}{c}{\text{Treatment class}} & & \\
        \addlinespace
        W=0 & W=1 & & \\
        \addlinespace
        \begin{block}{[cc]ccc}
             \topstrut \pi_0^T F_0^T(\dot{\phi}) & \pi_0^T\big(1-F_0^T(\dot{\phi})\big) & $Y=0$ & \multirow{2}{*}{Outcome class}\\
            \addlinespace
             \botstrut \pi_1^T F_1^T(\dot{\phi}) & \pi_1^T\big(1-F_1^T(\dot{\phi})\big) & $Y=1$\\
      \end{block}
    \end{blockarray}
\end{equation}
Likewise, the control sample causal confusion matrix is denoted by $\dot{\mathbf{CF}}^C$ and defined as:
\begin{equation} \label{eq:POC}
\dot{\mathbf{CF}}^C := \hspace{0.5cm}
    \begin{blockarray}{ccccc}
        \BAmulticolumn{2}{c}{\text{Treatment class}} & & \\
        \addlinespace
        W=0 & W=1 & & \\
        \addlinespace
        \begin{block}{[cc]ccc}
             \topstrut \pi_0^C F_0^C(\dot{\phi}) & \pi_0^C\big(1-F_0^C(\dot{\phi})\big) & $Y=0$ & \multirow{2}{*}{Outcome class}\\
            \addlinespace
             \botstrut \pi_1^C F_1^C(\dot{\phi}) & \pi_1^C\big(1-F_1^C(\dot{\phi})\big) & $Y=1$\\
      \end{block}
    \end{blockarray}
\end{equation}
The sample causal confusion matrix strongly resembles the confusion matrix, as defined in Section \ref{sec:costsensclas}, but contrasts the treatment class with the outcome class instead of contrasting the prediction class with the outcome class. Note that the positive treatment was applied to all instances in the treatment sample, including the instances that are causally classified in the negative treatment class. Similarly, the negative treatment was applied to instances in the positive treatment class in the control sample. 

\subsection{Performance measures}

A number of measures that have been proposed in the literature to assess the performance of causal classification models can be defined in terms of the sample causal confusion matrices of the control and treatment sample. To this end, we first define the positive treatment rate, $\dot{\eta}$, which is the causal equivalent of the positive prediction rate, $\eta$.

\begin{definition}
    The \textbf{positive treatment rate} is the proportion of instances in  $\mathcal{D}=\{\mathcal{D}^T,\mathcal{D}^C\}$ with a score above the causal classification threshold $\dot \phi$:
\begin{equation}\label{eq:actionrate}
    \dot{\eta} := \frac{\Big(\pi_0^C \big(1-F_0^C(\dot{\phi})\big) + \pi_1^C \big(1-F_1^C(\dot{\phi})\big)\Big) + \Big(\pi_0^T \big(1-F_0^T(\dot{\phi})\big) + \pi_1^T \big(1-F_1^T(\dot{\phi})\big)\Big)}{2}.
\end{equation}
\end{definition}

The positive treatment rate is a function of the causal classification threshold. The inverse function yields the equivalent causal classification threshold of the positive treatment rate, i.e.,  $\dot{\eta}^{-1}=\dot{\phi}$. Note that the positive treatment rates within the treatment sample and control sample, which we denote by  $\dot{\eta}(\dot{\phi})^T$ and $\dot{\eta}(\dot{\phi})^C$ for $\mathcal{D}^T$ and $\mathcal{D}^C$, respectively, resulting from the causal classification threshold, $\dot \phi$, may be different from the overall positive treatment rate $\dot{\eta}(\dot{\phi})$ across both groups, i.e., in $\mathcal{D}$. 

The Qini curve \cite{radcliffe2007using}, $\text{Qini}(\dot{\eta})$, is a frequently used visual approach for representing causal classification model performance. It is the causal equivalent of the gains curve for classification and reports the cumulative increase in positive outcomes as a function of the positive treatment rate:
\begin{equation}
    \text{Qini}(\dot{\eta}) := \pi_1^T \big(1-F_1^T(\dot{\eta}^{-1})\big) - \pi_1^C \big(1-F_1^C(\dot{\eta}^{-1})\big).
\end{equation}
Whereas the gains curve is a monotonically increasing function, the Qini curve may not be. Since the individual treatment effect can be positive as well as negative, the increase in positive outcomes for a positive treatment rate below $100\%$ may be greater than that obtained for a larger positive treatment rate, e.g., equal to $100\%$. The net increase in positive outcomes equals the difference between positive converted instances (with $ITE=1$) and negative converted instances (with $ITE=-1$). Hence, applying the treatment to all subjects may yield a smaller net increase in positive outcomes than applying the treatment to a subset of subjects.  

The Qini coefficient, $Q$, is the causal counterpart of the Gini coefficient \cite{radcliffe2007using}. It is equal to the ratio of the area between the Qini curves of the model and the random model, and the area between the Qini curves of a perfect model and the random model. A random model hence achieves a Qini coefficient of zero.
\begin{equation}\label{eq:Qini}
    Q := \frac{\int \text{Qini}(\dot{\eta}) \text{d} \dot{\eta} - \frac{\pi_1^T-\pi_1^C}{2}}{\frac{\pi_1^T-{\pi_1^C}^2-{\pi_1^T}^2}{2}}.
\end{equation}
The perfect model is defined as achieving the maximum increase in positive outcomes, which is equal to the positive outcome rate in the treatment sample, for a positive treatment rate that is equal to the positive outcome rate in the treatment sample \cite{radcliffe2007using}. When defining the Qini curve of the perfect model as such, it is assumed that all positive outcomes in the treatment sample are caused by the treatment and would have been negative if the treatment had not been applied, and that all positive instances in the control sample would have been negative if the treatment were applied. These assumptions regarding the perfect model ensure that the Qini coefficient is between zero and one but are unrealistic and therefore negatively affect the interpretability of the Qini coefficient. 

As an alternative approach, potential negative treatment effects may be ignored in defining the Qini curve of the perfect model, which as a result is monotonically increasing. The resulting coefficient, not bounded above by one, is called the little Qini coefficient, denoted by $q_0$ \cite{radcliffe2007using}:
\begin{align}\label{eq:littleQini}
    q_0 &:= \frac{\int \text{Qini}(\dot{\eta}) \text{d} \dot{\eta}-\frac{\pi_1^T-\pi_1^C}{2}}{\frac{\pi_1^T-\pi_1^C}{2}-\frac{(\pi_1^T-\pi_1^C)^2}{2}}, \nonumber \\
    &= \frac{2\int \frac{\text{Qini}(\dot{\eta})}{\pi_1^T-\pi_1^C}\text{d}\dot{\eta} - 1}{1 - (\pi_1^T-\pi_1^C)}.
\end{align}

The liftup curve, $\dot \lambda (\dot{\eta})$, is the causal counterpart of the lift curve \cite{devriendt2019} and plots the ratio of the increase in positive outcomes for positive treatment rate $\dot{\eta}$ and the increase in positive outcomes for $\dot{\eta}=1$, as a function of $\dot{\eta}$:
\begin{align}\label{eq:Liftup}
    \dot \lambda(\dot{\eta}) :&= \frac{\pi_1^T \big(1-F_1^T(\dot{\eta}^{-1})\big) - \pi_1^C \big(1-F_1^C(\dot{\eta}^{-1})\big)}{\dot{\eta} (\pi_1^T-\pi_1^C)}, \nonumber \\
    &= \frac{\text{Qini}(\dot\eta)}{(\pi_1^T - \pi_1^C) \dot\eta}.
\end{align}

\section{Cost-sensitive causal classification performance}\label{sec:costsenscausal}

\subsection{Profit}\label{subsec:profit}
In this section, we develop a framework for evaluating causal classification models in terms of the profit that would result from applying the model. The need for such a framework stems from its potential use (1) for evaluating the performance of causal classification models, (2) for optimizing the causal classification threshold \cite{elkan2001foundations} and (3) for defining an objective function and learning a causal classification model from data in line with the business objective \cite{zadrozny2001learning}.

To report the outcome distribution that would be observed for the positive and negative treatment class at causal classification threshold $\dot \phi$, i.e., when the prescribed treatment would have been applied, we introduce the causal confusion matrix, $\dot{\mathbf{CF}}$. The causal confusion matrix is the causal counterpart of the confusion matrix for conventional classification; it is equivalent to the treatment-response matrix as introduced by \cite{kane2014mining}.

\begin{definition}
The \textbf{causal confusion matrix}, $\dot{\mathbf{CF}}$, reports the proportion of negative and positive outcomes in the negative treatment class of the control sample, and the proportion of negative and positive outcomes in the positive treatment class of the treatment sample, for the causal classification threshold $\dot \phi$. 
\begin{equation} \label{eq:AO}
\dot{\mathbf{CF}} := \hspace{0.5cm}
    \begin{blockarray}{ccccc}
        \BAmulticolumn{2}{c}{\text{Treatment class}} & & \\
        \addlinespace
        W=0 & W=1 & & \\
        \addlinespace
        \begin{block}{[cc]ccc}
             \topstrut \pi_0^C F_0^C(\dot \phi) & \pi_0^T\big(1-F_0^T(\dot \phi)\big) & $Y=0$ & \multirow{2}{*}{Outcome class}\\
            \addlinespace
             \botstrut \pi_1^C F_1^C(\dot \phi) & \pi_1^T\big(1-F_1^T(\dot \phi)\big) & $Y=1$\\
      \end{block}
    \end{blockarray}
\end{equation}
\end{definition}

\begin{proposition}\label{propositionAO}
    $\Sigma_{i,j}(\dot{\mathbf{CF}})=1$
\end{proposition}
The proof of Proposition \ref{propositionAO} is provided in the Appendix. To caculate the profit that is associated with the outcomes as reported in the causal confusion matrix, we introduce two matrices: 

\begin{definition}
The \textbf{outcome-benefit matrix}, $\mathbf{OB}$, defines the benefit of an outcome depending on the treatment that is applied. The benefit of obtaining outcome $y_i$ when applying treatment $w_j$ is denoted by $b_{ij}$.
\begin{equation} \label{eq:OB}
\mathbf{OB} := \hspace{0.5cm}
    \begin{blockarray}{ccccc}
        \BAmulticolumn{2}{c}{\text{Treatment class}} & & \\
        \addlinespace
        W=0 & W=1 & & \\
        \addlinespace
        \begin{block}{[cc]ccc}
             \topstrut b_{00} & b_{01} & $Y=0$ & \multirow{2}{*}{Outcome class}\\
            \addlinespace
             \botstrut b_{10} & b_{11} & $Y=1$\\
      \end{block}
    \end{blockarray}
\end{equation}
\end{definition}

\begin{definition}
The \textbf{treatment-cost matrix}, $\mathbf{TC}$, defines the cost of a treatment depending on the outcome that is obtained. The cost of applying treatment $w_j$ when obtaining outcome $y_j$ is denoted by $c_{ij}$.
\begin{equation} \label{eq:AC}
\mathbf{TC} := \hspace{0.5cm}
    \begin{blockarray}{ccccc}
        \BAmulticolumn{2}{c}{\text{Treatment class}} & & \\
        \addlinespace
        W=0 & W=1 & & \\
        \addlinespace
        \begin{block}{[cc]ccc}
             \topstrut c_{00} & c_{01} & $Y=0$ & \multirow{2}{*}{Outcome class}\\
            \addlinespace
             \botstrut c_{10} & c_{11} & $Y=1$\\
      \end{block}
    \end{blockarray}
\end{equation}
\end{definition}
By convention, the benefit and cost values in the outcome-benefit and treatment-cost matrix are positive or equal to zero. The $\mathbf{OB}$ and $\mathbf{TC}$ matrices allow calculating the causal cost-benefit matrix, $\mathbf{\dot{CB}}$.

\begin{definition}
The \textbf{causal cost-benefit matrix}, $\mathbf{\dot{CB}}$, defines the profit of an outcome depending on the treatment that is applied and can be calculated from the outcome-benefit and treatment-cost matrix as follows:
\begin{equation}
\mathbf{\dot{CB}}=\mathbf{OB} - \mathbf{TC}.
\label{eq:CCB}
\end{equation}
\end{definition}
The values in the $\mathbf{\dot{CB}}$ matrix are positive for treatment-outcome combinations that yield a net profit, and values will be negative for combinations that yield a net loss. The $\mathbf{\dot{CB}}$ matrix allows defining the profit per instance of a causal classification model. 

\begin{definition} The \textbf{profit per instance}, $P$, of  a causal classification model $\dot m$ with causal classification threshold $\dot{\phi}$ equals the difference between the benefit per instance resulting from the obtained outcomes and the cost per instance resulting from the applied treatments:
\begin{align}\label{eq:def_UP}
    P &= \Sigma_{i,j}(\dot{\mathbf{CF}}_{ij} \circ \mathbf{OB}_{ij} - \dot{\mathbf{CF}}_{ij} \circ \mathbf{TC}_{ij}), \nonumber\\ 
    &= \Sigma_{i,j}\big(\dot{\mathbf{CF}}_{ij} \circ (\mathbf{OB}_{ij} - \mathbf{TC}_{ij})\big), \nonumber\\ 
    &= \Sigma_{i,j}(\dot{\mathbf{CF}}_{ij} \circ \mathbf{\dot{CB}}_{ij})
\end{align}
\end{definition}

Note the strong resemblance between Equation \eqref{eq:profClass} and Equation \eqref{eq:def_UP}, with the causal confusion matrix being the causal counterpart of the confusion matrix and the causal cost-benefit matrix being the causal counterpart of the cost-benefit matrix. Using the notation introduced above, we obtain the following expression for the profit per instance, $P$:
\begin{align}
    P = &\pi_0^C F_0^C(\dot{\phi})(b_{00}-c_{00}) + \pi_1^C F_1^C(\dot{\phi})(b_{10}-c_{10})\\ \nonumber     &+\pi_0^T\big(1-F_0^T(\dot{\phi})\big) (b_{01}-c_{01}) + \pi_1^T\big(1-F_1^T(\dot{\phi})\big) (b_{11}-c_{11}).  \nonumber 
\end{align}

\subsection{Causal profit}\label{subsec:causalprofit}

The profit per instance defined in Equation \eqref{eq:def_UP} implicitly adopts a baseline that defines the meaning of the value zero; i.e., no treatments are applied and no outcomes are obtained. As such, the resulting profit is the absolute profit, as defined in Section \ref{sec:relperf}. In a causal classification problem context, however, this is not a meaningful baseline since the objective is to estimate the effect of the positive treatment on the outcome that occurs, relative to the baseline scenario of the outcome that occurs when the negative treatment is applied, in order to optimize the selection of instances that are to be treated with the positive treatment. Hence, when adopting a cost-sensitive perspective in assessing the performance of a causal classification model, the net increase in profit is to be calculated relative to the baseline scenario of applying the negative treatment to all instances, which we henceforth will call the negative treatment baseline scenario. 

In the negative treatment baseline scenario, the negative treatment class comprises the entire population, and the baseline outcome distribution is indicated by the baseline causal confusion matrix, $\mathbf{\dot{CF}_b}$:
\begin{equation}\label{eq:AO0}
    \mathbf{\dot{CF}_b} = 
\begin{bmatrix}
\pi_0^C & 0\\
\pi_1^C & 0
\end{bmatrix}.
\end{equation}
Note that, by definition, $\mathbf{\dot{CF}_b} = \dot{\mathbf{CF}}^C$ for $\dot{\phi}>max(t)$. 

\begin{definition}
The \textbf{causal effect matrix}, $\mathbf{\dot{E}}$, reports the shift in the outcome distribution that results from applying the positive treatment $W=1$ to the instances in the positive treatment class, compared to the outcome distribution in the negative treatment baseline scenario. The causal effect matrix is calculated as the difference between the causal confusion matrix and the causal confusion matrix of the negative treatment baseline scenario, which we denote by $\mathbf{\dot{CF}_b}$. 
\begin{equation}\label{eq:cematrix}
    \mathbf{\dot{E}} = \dot{\mathbf{CF}} - \mathbf{\dot{CF}_b} 
\end{equation}
\end{definition}

The causal effect matrix is the causal equivalent of the effect matrix as defined in Equation \eqref{eq:E} for conventional classification. Elaborating Equation \eqref{eq:cematrix} by entering Equation \eqref{eq:AO} and \eqref{eq:AO0} of the causal confusion matrices $\dot{\mathbf{CF}}$ and $\dot{\mathbf{CF_b}}$, respectively, where $\pi_0^C F_0^C(\dot{\phi}) - \pi_0^C = - \pi_0^C\big(1-F_0^C(\dot{\phi})\big)$ and $\pi_1^C F_1^C(\dot{\phi}) - \pi_1^C = - \pi_1^C\big(1-F_1^C(\dot{\phi})\big)$, yields:
\begin{equation} \label{eq:CE}
\mathbf{\dot{E}} := \hspace{0.5cm}
    \begin{blockarray}{ccccc}
        \BAmulticolumn{2}{c}{\text{Treatment class}} & & \\
        \addlinespace
        W=0 & W=1 & & \\
        \addlinespace
        \begin{block}{[cc]ccc}
             \topstrut - \pi_0^C\big(1-F_0^C(\dot{\phi})\big) & \pi_0^T \big(1-F_0^T(\dot{\phi})\big) & $Y=0$ & \multirow{2}{*}{Outcome class}\\
            \addlinespace
             \botstrut - \pi_1^C\big(1-F_1^C(\dot{\phi})\big) & \pi_1^T \big(1-F_1^T(\dot{\phi})\big) & $Y=1$\\
      \end{block}
    \end{blockarray}
\end{equation}

\begin{proposition}\label{propositionCE}
$\Sigma_{i,j}(\mathbf{\dot{E}})=0$
\end{proposition}

The proof of Proposition \ref{propositionCE} is provided in the Appendix. Intuitively, we expect that the aggregated decrease in the number of instances of the positive and negative classes in the negative treatment class effectively equals the aggregated increase in the number of instances of the positive and negative classes in the positive treatment class. 

\begin{definition}
The \textbf{causal profit per instance}, $\dot{P}$, of a causal classification model $\dot m$ with causal classification threshold $\dot{\phi}$ equals the difference between the profit per instance, $P$, as defined in Equation \eqref{eq:def_UP}, and the profit per instance that is obtained in the negative treatment baseline scenario, which we denote by $P_b$:  

\begin{align}\label{eq:CP1}
    \dot{P} := P - P_{b} 
\end{align}
\end{definition}

The relation between the causal profit, the causal effect matrix and the profit matrix is established by the following proposition:

\begin{proposition}\label{propositionPiC}
$\dot{P} = \Sigma_{i,j}(\mathbf{\dot{E}}_{ij} \circ \mathbf{\dot{CB}}_{ij})$
\end{proposition}

\noindent The proof of Proposition \ref{propositionPiC} is provided in the Appendix. The definition of causal profit in Equation \eqref{eq:CP1} is equivalent with the definition of relative profit in Equation \eqref{eq:baseline} for conventional classification, with a baseline model that prescribes a baseline treatment, i.e., the negative treatment, to be applied to all instances. By elaborating Equation \eqref{eq:CP1} using Proposition \ref{propositionPiC}, we obtain the following formula to calculate the causal profit, $\dot{P}$:

\begin{align}\label{eq:CP2}
    \dot{P} = & - \pi_0^C \big(1-F_0^C(\dot{\phi})\big)(b_{00}-c_{00}) + \pi_0^T\big(1-F_0^T(\dot{\phi})\big) (b_{01}-c_{01})\nonumber \\
    & - \pi_1^C \big(1-F_1^C(\dot{\phi})\big)(b_{10}-c_{10}) + \pi_1^T\big(1-F_1^T(\dot{\phi})\big) (b_{11}-c_{11}).
\end{align}

\subsection{Maximum Causal Profit and Expected Maximum Causal Profit}\label{subsec:mcp}

The profit as calculated following Equation \eqref{eq:CP2} is a function of the causal classification threshold $\dot{\phi}$. Extending upon the maximum profit measure for classification, we propose to maximize the causal profit per instance as expressed by Equation \eqref{eq:CP2} by optimizing the threshold $\dot{\phi}$. The resulting maximum causal profit per instance, $\dot{MP}$, is a business-oriented measure for evaluating the performance of a causal classification model. 

\begin{definition}
The\textbf{maximum causal profit} measure, $\dot{MP}$, for evaluating a causal classification model is defined as follows::
\begin{align}\label{eq:MCP}
    \dot{MP} &:= \max_{\forall \dot{\phi}} \big(\dot{P}(\dot{\phi};\dot{\mathbf{CB}})\big) \\ \nonumber
    &= \max_{\forall \dot{\phi}} \big( \Sigma_{i,j}(\mathbf{\dot{E}}_{ij} \circ \mathbf{\dot{CB}}_{ij}) \big) \\ \nonumber
    &= \max_{\forall \dot{\phi}} \Big( \Sigma_{i,j} \big(\mathbf{\dot{E}}_{ij} \circ (\mathbf{OB}_{ij} - \mathbf{TC}_{ij}) \big) \Big) \\ \nonumber
    &= \dot{P}(\dot{\phi}^{*}; \dot{\mathbf{CB}}) \\ \nonumber
\end{align}
\end{definition}

\begin{definition}
The \textbf{causal profit maximizing positive treatment rate}, $\eta_{\dot{MP}}$, is the proportion of instances to treat with the positive treatment in order to maximize the profit of the intervention:
\begin{equation}
    \eta_{\dot{MP}} = \pi_0^T \big(1-F_0^T(\dot{\phi}^{*})\big)+\pi_1^T \big(1-F_1^T(\dot{\phi}^{*})\big)
\end{equation}
\end{definition}

The maximum causal profit measure is an intuitively interpretable measure that allows the evaluation and comparison of causal classification models in terms of their potential for maximizing the business outcome, i.e., profit. The $\dot{MP}$ measure may complement the use of cost-insensitive performance measures, such as the Qini, liftup or AUCROC measures. Moreover, $\dot{MP}$ supports practitioners in decision-making by providing the optimal causal classification threshold, $\dot{\phi}^{*}$, and the associated causal profit maximizing positive treatment rate for segmenting the population into a positive and negative treatment class based on the ITE estimates of the causal classification model. 

In a similar manner as the MP measure for classification was extended to the EMP measure, we extend upon the $\dot{MP}$ measure and define the expected maximum causal profit measure.
\begin{definition}
The \textbf{expected maximum causal profit} measure, $\dot{EMP}$, for evaluating a causal classification model is defined as follows:
\begin{equation}\label{eq:EMCP}
    \dot{EMP} := \int_{-\infty}^{\infty} \dot{P}(\dot{\phi}^{*},\dot{\mathbf{CB}}) \cdot h(\dot{\mathbf{CB}}) d\dot{\mathbf{CB}}
\end{equation}
with $h(\dot{\mathbf{CB}})$ the joint distribution of the parameters of the causal cost-benefit matrix. 
\end{definition}

The main advantages of the $\dot{EMP}$ measure are its robustness and the flexibility that it provides to the user in adapting to the applicable operating conditions. Further extensions are possible in line with \cite{orallo2012} to address the uncertainty regarding the operating conditions at the time of model evaluation. The primary disadvantages of the $\dot{EMP}$ measure are the difficulty in specifying an appropriate joint probability distribution for the cost-benefit parameters and the added complexity. However, in practical applications, several cost and benefit parameters may take a zero value and a limited number of nonzero parameters may be uncertain or random, simplifying the practical use of the $\dot{EMP}$ measure.

The interpretation of the measure remains the same as for the $\dot{MP}$ measure. Whereas the $\dot{MP}$ measure relies upon the representativeness of the sample and the ability of the user to specify an accurate value for the cost and benefit parameters, the $\dot{EMP}$ measure provides the means to achieve a measure that better reflects the \textit{true} average causal profit that will be obtained when applying the causal classification model under a range of operating conditions. 

When comparing, on the one hand, Equations \eqref{eq:MCP} and \eqref{eq:EMCP} of the $\dot{MP}$ and $\dot{EMP}$, respectively, and on the other hand Equations \eqref{eq:MP} and \eqref{eq:EMP} of the MP and EMP, respectively, we find that the causal counterpart of the confusion matrix is the causal confusion matrix, of the effect matrix the causal effect matrix, and of the cost-benefit matrix the causal cost-benefit matrix. Moreover, extending Proposition \ref{propositionClassification}, we can prove that conventional classification concerns a specific case of causal classification from a cost-sensitive evaluation perspective. Proposition \ref{propositionClassificationCostSens} below extends Proposition \ref{propositionClassification} and is to be interpreted in the same manner. 

\begin{proposition}\label{propositionClassificationCostSens}
Causal classification as defined in terms of the causal profit, maximum causal profit and expected maximum causal profit measures instantiates to conventional classification as defined in terms of the profit, maximum profit and expected maximum profit measures when the number of treatments that can be applied is exactly equal to one, i.e., $\lvert W \rvert = 1$.
\end{proposition}

\noindent \textbf{Proof of Proposition \ref{propositionClassificationCostSens}}
If a single treatment can be applied, $W \in \{1\}$, then treatment $W=1$  is prescribed for both the instances in the positive and negative treatment class, as defined by the classification threshold $\dot{\phi}$. \\
As a result, the causal confusion matrix, $\dot{\mathbf{CF}}$, as defined in Equation \eqref{eq:AO}, instantiates to the confusion matrix, $\mathbf{CF}$, as defined in Equation \eqref{eq:CF}.\\
Hence, the formula of the causal profit, maximum causal profit and expected maximum causal profit, instantiate to the formula of profit, maximum profit and expected maximum profit for classification, respectively, with the causal cost-benefit matrix, $\dot{\mathbf{CB}}$, as defined in Equation \eqref{eq:CCB} instantiating to the classification cost-benefit matrix, $\mathbf{CB}$, as defined in Equation \eqref{eq:CB}.\\
\QEDB

\section{Practical applications}\label{sec:instantiation}

In this section, we instantiate the causal profit measure to evaluate causal classifiers for customer retention and customer response modeling, as previously proposed in the literature\cite{devriendt2019,gubela2019response}. As such, we indirectly demonstrate the practical use of the framework for evaluating causal classification models. 

\subsection{Customer retention}

\cite{devriendt2019} apply causal classification methods for developing customer churn uplift models, allowing estimation if the effect of a retention campaign on the churn risk of an individual customer. This approach allows optimization of the selection of customers intended to be targeted with the retention campaign. \cite{devriendt2019} propose the following formula to calculate the total profit of a retention campaign (Equation 12, page 18):
\begin{align}\label{eq:profitDevriendt}
    P^{\text{Retention}} = N \alpha \big((\beta_C-\beta_T)(CLV-C_c-C_i) - (1-\beta_C)(C_c+C_i)- \beta_T C_c\big),
\end{align}
\noindent where $N$ is the total number of customers and $\alpha$ is the proportion of customers that is targeted; hence, $N \alpha$ is the number of customers in the positive treatment class; $\beta_C$ and $\beta_T$ are the churn rates in the control and treatment sample positive treatment class, respectively; $CLV$ is the customer lifetime value, i.e., the benefit of a customer who does not churn; and $C_c$ is the cost of contacting a customer and $C_i$ is the cost of the incentive that is offered in the retention campaign.

The outcome of interest in this application is whether a customer will churn ($Y=0$) or not ($Y=1$). A decision is to be made whether to target a customer with a retention campaign ($W=1$) or not ($W=0$), offering an incentive to remain loyal, e.g., a discount. The retention campaign is the positive treatment that may alter the outcome. 

Table \ref{tab:cost_benefit_churn}a shows the applicable outcome-benefit matrix. The benefit of a customer who does not churn, i.e., the benefit of a positive outcome, is independent of the treatment that is applied and equal to the customer lifetime value. The $CLV$ is the average net discounted future cash flow that is expected to be generated by a customer over a predefined time window. Table \ref{tab:cost_benefit_churn}b shows the applicable treatment-cost matrix. Targeting a customer with a retention campaign involves a cost of contacting the customer, $C_c$. If the customer accepts the retention offer, an additional cost is incurred, i.e., the cost of the incentive, $C_i$. Note that the cost of the incentive is only incurred when the customer does not churn, i.e., for a positive outcome. In this application, the negative treatment implies not targeting a customer and does not involve a cost. The outcome-benefit and treatment-cost matrices align with the benefits and costs in the profit formula proposed in \cite{neslin2006defection}, which underlies both the maximum profit measure for churn \cite{verbeke2012new} and the $MPU$ measure. 
\begin{table}[ht]
    \centering
    \caption{Outcome-benefit and treatment-cost matrices for customer churn uplift modeling.}
	\begin{adjustbox}{width=0.55\textwidth}
	\begin{tabular}{l l c c}
	\noalign{\smallskip}\noalign{\smallskip}\noalign{\smallskip}
	& &\multicolumn{2}{c}{Treatment} \\
	\noalign{\smallskip}\noalign{\smallskip}
	& & W=0 & W=1\\
	\noalign{\smallskip}\hline\noalign{\smallskip}
	\multirow{2}{*}{{\rotatebox[origin=c]{90}{\small{Outcome}}}}  & \multirow{1}{*}{Y=0} & 0 & 0\\
	\noalign{\smallskip}\cline{2-4}\noalign{\smallskip}
	& \multirow{1}{*}{Y=1} & $CLV$ & $CLV$\\
	\noalign{\smallskip}\hline
	\noalign{\smallskip}
	\multicolumn{4}{c}{(a) OB matrix}\\
	\end{tabular}
	\hspace{2em} 
	\begin{tabular}{l l c c}
	\noalign{\smallskip}\noalign{\smallskip}\noalign{\smallskip}
	& &\multicolumn{2}{c}{Treatment} \\
	\noalign{\smallskip}\noalign{\smallskip}
	& & W=0 & W=1\\
	\noalign{\smallskip}\hline\noalign{\smallskip}
	\multirow{2}{*}{{\rotatebox[origin=c]{90}{\small{Outcome}}}}  & \multirow{1}{*}{Y=0} & 0 & $C_c$\\
	\noalign{\smallskip}\cline{2-4}\noalign{\smallskip}
	& \multirow{1}{*}{Y=1} & 0 & $C_c + C_i$\\
	\noalign{\smallskip}\hline
	\noalign{\smallskip}
	\multicolumn{4}{c}{(b) TC matrix}\\
	\end{tabular}
	\end{adjustbox}
	 \label{tab:cost_benefit_churn}
\end{table}

Given the outcome-benefit and treatment-cost matrix of Table \ref{tab:cost_benefit_churn}, we can elaborate Equation \eqref{eq:CP2} for customer retention as follows:
\begin{align}\label{eq:CPChurn}
    \dot{P} = - \pi_0^T \big(1-F_0^T(\dot{\phi})\big) C_c - \pi_1^C \big(1-F_1^C(\dot{\phi})\big) CLV + \pi_1^T \big(1-F_1^T(\dot{\phi})\big) (CLV - C_c - C_i).
\end{align}

Note that $\beta_T$ and $\beta_C$ in Equation \eqref{eq:profitDevriendt} denote the churn rate in the positive treatment class of the treatment and the control sample, respectively, and $(1-\beta_T)$ and $(1-\beta_C)$ indicate the proportion of non-churners in the positive treatment class of the treatment and the control sample, respectively. 

Using the following equalities:
\begin{align} 
\pi_0^T \big(1-F_0^T(\dot{\phi})\big) &= \alpha \beta_T,\\
\pi_1^C \big(1-F_1^C(\dot{\phi})\big) &= \alpha(1 - \beta_C), \\
\pi_1^T \big(1-F_1^T(\dot{\phi})\big) &= \alpha(1 - \beta_T), \nonumber \\
    &= \alpha(1-\beta_C) + \alpha(\beta_C-\beta_T),
\end{align}

we can rework the causal profit per instance formula of Equation \eqref{eq:CPChurn} to arrive at Equation \eqref{eq:profitDevriendt}:
\begin{align}
   \dot{P} &= \big(\alpha(1-\beta_C) + \alpha(\beta_C-\beta_T)\big) (CLV - C_c - C_i) - \alpha(1 - \beta_C) CLV - \alpha \beta_T C_c, \nonumber \\
    &= \alpha(\beta_C-\beta_T)\big) (CLV - C_c - C_i) + \alpha(1-\beta_C) (CLV - C_c - C_i) - \alpha(1 - \beta_C) CLV - \alpha \beta_T C_c, \nonumber \\
    &= \alpha(\beta_C-\beta_T)\big) (CLV - C_c - C_i) + \alpha(1-\beta_C) CLV - \alpha(1-\beta_C)(C_c + C_i) - \alpha(1 - \beta_C) CLV - \alpha \beta_T C_c, \nonumber \\
    &= \alpha(\beta_C-\beta_T)\big) (CLV - C_c - C_i) - \alpha(1-\beta_C)(C_c + C_i) - \alpha \beta_T C_c, \nonumber \\
    &= \frac{P^{\text{Retention}}}{N}
\end{align}

Equation \eqref{eq:profitDevriendt} can be instantiated from the generic causal profit formula of Equation \eqref{eq:CP1}. The maximum profit uplift ($MPU$) measure, which is introduced in \cite{devriendt2019} and maximizes the profit calculated following Equation \eqref{eq:profitDevriendt}, hence is an instantiation of the general $\dot{MP}$ measure. 

\subsection{Customer response}

\cite{gubela2019response} apply causal classification methods for developing customer response models, allowing estimation of the incremental effect of a discount on the propensity to purchase of an individual customer. This approach allows optimization of the selection of customers intended to be targeted with the marketing campaign. \cite{gubela2019response} propose the following formula to calculate the total profit of a response campaign, $P^{\text{Response}}$ (Equation 13, page 653):
\begin{equation}\label{eq:profitresponse}
  P^{\text{Response}} = N_{\tau} ( \pi_{\tau} \delta_{\tau} - \pi_{\zeta} \delta_{\zeta}) - N_{\tau} \varepsilon_{unit} - \rho \sum_{i=1}^{N_{\tau}}\delta_{\pi_+,i},
\end{equation}
where $N_{\tau}$ is the number of instances in the positive treatment class, $N$ is equal to the total number of customers and $\tau$ is the proportion of customers that is targeted; $\pi_{\tau}$ is the proportion of positive outcomes in the positive treatment class, $\delta_{\tau}$ the average revenue in the positive treatment class, and the equivalent quantities as measured for a control sample are $\pi_{\zeta}$ and $\delta_{\zeta}$, respectively; $\varepsilon_{unit}$ is the cost of targeting a customer with the marketing campaign; and $\rho$ is the percentage discount  that  is given, $\pi_+$ is the proportion of responders in the positive treatment class, and $\delta_{\pi_+,i}$ the revenue generated by customer $i$ in the positive treatment class.

Equation \eqref{eq:profitresponse} can be instantiated from the generic causal profit formula of Equation \eqref{eq:CP2}. The outcome of interest is whether a customer makes a purchase ($Y=1$) or not ($Y=0$). The decision to be made is whether to target a customer with a response campaign ($W=1$) or not ($W=0$), e.g., offering a discount on a purchase. The response campaign is the treatment that may alter the outcome. 
\begin{table}[ht]
    \centering
    \caption{Outcome-benefit and treatment-cost matrices for customer response uplift modeling.}
	\begin{adjustbox}{width=0.55\textwidth}
	\begin{tabular}{l l c c}
	\noalign{\smallskip}\noalign{\smallskip}\noalign{\smallskip}
	& &\multicolumn{2}{c}{Treatment} \\
	\noalign{\smallskip}\noalign{\smallskip}
	& & W=0 & W=1\\
	\noalign{\smallskip}\hline\noalign{\smallskip}
	\multirow{2}{*}{{\rotatebox[origin=c]{90}{\small{Outcome}}}}  & \multirow{1}{*}{Y=0} & 0  & 0 \\
	\noalign{\smallskip}\cline{2-4}\noalign{\smallskip}
	& \multirow{1}{*}{Y=1} & $\delta^C$ & $\delta^T$\\
	\noalign{\smallskip}\hline
	\noalign{\smallskip}
	\multicolumn{4}{c}{(a) OB matrix}\\
	\end{tabular}
	\hspace{2em} 
	\begin{tabular}{l l c c}
	\noalign{\smallskip}\noalign{\smallskip}\noalign{\smallskip}
	& &\multicolumn{2}{c}{Treatment} \\
	\noalign{\smallskip}\noalign{\smallskip}
	& & W=0 & W=1\\
	\noalign{\smallskip}\hline\noalign{\smallskip}
	\multirow{2}{*}{{\rotatebox[origin=c]{90}{\small{Outcome}}}}  & \multirow{1}{*}{Y=0} & 0 & $C_c$\\
	\noalign{\smallskip}\cline{2-4}\noalign{\smallskip}
	& \multirow{1}{*}{Y=1} & 0 & $C_c + C_i$\\
	\noalign{\smallskip}\hline
	\noalign{\smallskip}
	\multicolumn{4}{c}{(b) TC matrix}\\
	\end{tabular}
	\end{adjustbox}
	 \label{tab:cost_benefit_marketing}
\end{table}

Table \ref{tab:cost_benefit_marketing} shows the outcome-benefit matrix and the treatment-cost matrix for this case. Note that the benefit of a positive outcome, here, depends on the treatment. 
We denote the constant cost of targeting a customer with $C_c = \varepsilon_{unit}$, and the average cost of the incentive with $C_i$. 
\cite{gubela2019response} consider a varying incentive cost. That is, the financial value of the promotional action varies according to the size of the discount and the shopping basket. In Equation \eqref{eq:CP2} for calculating the total profit, the total cost of the incentives is calculated. To calculate the profit per instance, we will take the average cost of the incentive into account, which is calculated as the ratio of the total cost of the incentives and the number of responders in the positive treatment class, $N_{\tau} \pi_{\tau}$:
\begin{equation}
    C_i = \frac{\rho \sum_{i=1}^{N_{\tau}}\delta_{\pi_+,i}}{N_{\tau} \pi_{\tau}}
\end{equation}
Using the following equalities,
\begin{align*}
    \tau \pi_{\tau} &= \pi_1^T\big(1-F_1^T(\dot{\phi})\big), \nonumber \\
    \tau \pi_{\zeta} &= \pi_1^C\big(1-F_1^C(\dot{\phi})\big), \nonumber \\
    \tau &= \pi_1^T\big(1-F_1^T(\dot{\phi})\big) + \pi_0^T\big(1-F_0^T(\dot{\phi})\big), \nonumber \\
    \rho \sum_{i=1}^{N_{\tau}}\delta_{\pi_+,i} &= N \tau \pi_{\tau} C_i, \nonumber \\
    &= N \pi_1^T\big(1-F_1^T(\dot{\phi})\big) C_i, \nonumber \\
\end{align*}

\noindent we can rework Equation \eqref{eq:profitresponse} to arrive at the causal profit per instance formula of Equation \eqref{eq:CP1}:
\begin{align*}
    \frac{P^{\text{Response}}}{N} \nonumber =& \tau \pi_{\tau} \delta_{\tau} - \tau \pi_{\zeta} \delta_{\zeta} - \tau \varepsilon_{unit} - \frac{\rho \sum_{i=1}^{N_{\tau}}\delta_{\pi_+,i}}{N}, \nonumber \\
     =& \pi_1^T\big(1-F_1^T(\dot{\phi})\big) \delta_{\tau} - \pi_1^C\big(1-F_1^C(\dot{\phi})\big) \delta_{\zeta} \\
    &- \pi_1^T\big(1-F_1^T(\dot{\phi})\big) C_c - \pi_0^T\big(1-F_0^T(\dot{\phi})\big) C_c
    - \pi_1^T\big(1-F_1^T(\dot{\phi})\big) C_i \nonumber \\
    =& \pi_1^T\big(1-F_1^T(\dot{\phi})\big) (\delta_{\tau} - C_c - C_i) + \pi_1^C\big(1-F_1^C(\dot{\phi})\big) \delta_{\zeta} \\
    =& \Sigma_{i,j}\big(\mathbf{\dot{E}}_{ij} \circ (\mathbf{OB}_{ij}-\mathbf{TC}_{ij})\big), \\
    =& \Sigma_{i,j}(\mathbf{\dot{E}}_{ij} \circ \mathbf{\dot{CB}}_{ij}), \\
    =& \dot{P}
\end{align*}

\section{Conclusions and research agenda}\label{sec:conclusions}

Classification models are adopted across a variety of operational business processes to predict an outcome of interest based on historical data. These models allow to anticipate the future and optimize managerial decision-making, by making decisions based on the predictions of the model. To improve upon the performance of conventional classification models, cost-sensitive classification models can be adopted to account for the benefit of correct predictions and the cost of incorrect classifications. Alternatively, causal classification models can be adopted to account for the causal effect of the action under consideration, i.e., the treatment, on the outcome of interest. The objective of this article is to integrate cost-sensitive and causal classification to facilitate optimization of resource allocation and maximize intervention effectiveness in terms of net returns\cite{gupta2020maximizing}. To this end, we propose a unifying framework for cost-sensitive and cost-insensitive evaluations of both causal and conventional classification models. 

We first elaborate a framework for evaluating conventional classification model performance, which mainly consists of the confusion matrix, $\mathbf{CF}$, and the cost-benefit matrix, $\mathbf{CB}$. Additionally, we define the effect matrix, $\mathbf{E}$, to formalize the use of various baseline models in assessing performance in a relative manner, i.e., in comparison with a baseline model. The effect $\mathbf{E}$ matrix is equal to the difference between the confusion matrices of the model and the baseline model. A variety of existing, cost-sensitive and cost-insensitive performance measures, can be defined in terms of the $\mathbf{CF}$, $\mathbf{CB}$ and $\mathbf{E}$ matrix, e.g., AUC, Gini and Lift.

In a second step, we extend the framework for conventional classification toward causal classification. To this end, we formulate the sample causal confusion matrix, $\dot{\mathbf{CF}}^S$, the treatment-cost matrix, $\mathbf{TC}$, and the outcome-benefit matrix, $\mathbf{OB}$, which subsequently allow the causal confusion matrix $\dot{\mathbf{CF}}$, the causal cost-benefit matrix, $\mathbf{\dot{CB}}$, and the causal effect matrix, $\mathbf{E}$, to be defined. These matrices facilitate the formalization of a variety of existing, cost-insensitive performance measures for evaluating causal classification measures, i.e., Qini, little Qini and liftup. Additionally, a novel visual approach and associated performance measure for causal classification model evaluation are proposed, i.e., the CROC curve and AUCROC measure, as the causal counterparts of the ROC curve and the AUC measure for conventional classification, respectively. 

Based on the presented framework, we propose three novel cost-sensitive measures for evaluating causal classification models, i.e., causal profit, maximum causal profit and expected maximum causal profit. These measures take into account both the causal effect and the cost of a treatment, as well as the cost or benefit of an outcome, in assessing the performance of a causal classification model. This approach allows optimization of the causal classification threshold for causally classifying instances, with the objective of maximizing the profit that results from applying the positive treatment to the selected instances in the positive treatment class, based on the predicted individual treatment effect by the causal classification model. As such, we effectively integrate cost-sensitive and causal classification and present a practical approach that supports business decision-making. 

In addition, we prove that a range of causal classification performance measures instantiate to a range of equivalent performance measures for assessing the performance of conventional classification model. Hence, we prove that from an evaluation perspective, conventional classification can be regarded as a specific case of causal classification, i.e., with the number of possible treatments that can be applied equal to one. As such, the proposed framework unifies cost-insensitive and cost-sensitive, causal and conventional classification. 

The proposed framework stands on the shoulders of the cost-sensitive and causal learning paradigms. Their integration opens a range of opportunities for future research, such as the extension of the proposed cost-sensitive causal classification framework beyond the double-binary causal classification setting that was elaborated in this article. Topics of prime interest in this regard involve an extension of the framework to multitreatment effect estimation, to multivalued outcomes (similar to multiclass classification), to multiple outcomes, to continuous treatments, to continuous outcomes, and to instance-dependent (also called observation-dependent or example-dependent) costs and benefits. Another extension of interest concerns the cost of acquiring data through field experiments, which could be taken into account in adopting a broader perspective toward the use of causal classification.

\bibliographystyle{plain}
\bibliography{bibliography.bib}

\newpage

\section*{Appendix}

\noindent \textbf{Proof of Proposition \ref{proposition:E}}
\noindent By definition, the sum of the elements of a confusion matrix is equal to one, i.e., $\sum_{i,j}\mathbf{CF}=\sum_{i,j}\mathbf{CF_b}=1$.
Hence, we have: 
\begin{align*}
    \sum_{i,j}\mathbf{E}_{ij} &= \sum_{i,j}(\mathbf{CF}_{ij}-\mathbf{CF_b}_{,ij})\\
    &= \sum_{i,j}\mathbf{CF}_{ij}-\sum_{i,j}\mathbf{CF_b}_{,ij}\\
    &= 1-1 \\
    &= 0 
\end{align*}
\QEDB

\noindent \textbf{Proof of Proposition \ref{propositionPO}}
\begin{align*}
    \Sigma_{i,j}(\dot{\mathbf{CF}}_{i,j}^S) &= \pi_0^S F_0^S(\dot{\phi})  + \pi_0^S \big(1-F_0^S(\dot{\phi})\big) + \pi_1^S  F_1^S(\dot{\phi}) + \pi_1^S \big( 1-F_1^S(\dot{\phi})\big)\\
    &= \pi_0^S F_0^S(\dot{\phi}) - \pi_0^S F_0^S(\dot{\phi}) + \pi_0^S + \pi_1^S F_1^S(\dot{\phi}) - \pi_1^S F_1^S(\dot{\phi}) +  \pi_1^S \\
  &= \pi_0^S + \pi_1^S \\
  &= 1 
\end{align*}
\QEDB

\noindent \textbf{Proof of Proposition \ref{propositionAO}}
\begin{align*}
    \Sigma_{i,j}(\dot{\mathbf{CF}}) &= \pi_0^C F_0^C(\dot{\phi}) + \pi_1^T \big(1-F_1^T(\dot{\phi})\big) + \pi_1^C  F_1^C(\dot{\phi}) + \pi_0^T \big(1-F_0^T(\dot{\phi})\big) \\
    &= \pi_0^C F_0^C(\dot{\phi}) + \pi_1^T - \pi_1^T F_1^T(\dot{\phi}) + \pi_1^C F_1^C(\dot{\phi}) + \pi_0^T - \pi_0^T F_0^T(\dot{\phi}) \\
    &= 1 + \pi_0^C F_0^C(\dot{\phi}) + \pi_1^C F_1^C(\dot{\phi})- \pi_1^T F_1^T(\dot{\phi}) - \pi_0^T F_0^T(\dot{\phi}) \end{align*}

Hence, we need to prove that:

\begin{equation*}
\pi_0^C F_0^C(\dot{\phi}) + \pi_1^C F_1^C(\dot{\phi})- \pi_1^T F_1^T(\dot{\phi}) - \pi_0^T F_0^T(\dot{\phi}) = 0\\
\end{equation*}

Which is achieved when:

\begin{equation*}
\pi_0^C F_0^C(\dot{\phi}) + \pi_1^C F_1^C(\dot{\phi}) = \pi_1^T F_1^T(\dot{\phi}) + \pi_0^T F_0^T(\dot{\phi})
\end{equation*}

We define the cumulative distribution of treatment and control sample instances as a function of the causal model score $t$, which we denote by $F^T(t)$ and $F^C(t)$, respectively: 
\begin{align*}
F^T(t) &:= \pi_0^T F_0^C(t) + \pi_1^T F_1^T(t) \\
F^C(t) &:= \pi_0^C F_0^C(t) + \pi_1^C F_1^C(t)
\end{align*}

Hence, $\Sigma_{i,j}(\dot{\mathbf{CF}})=1$ if $F^T(t)=F^C(t)$, which asymptotically holds for increasing sample sizes when the treatment and control sample are independent and identically distributed random samples from the population.
\QEDB

\vspace{1cm}
\noindent \textbf{Proof of Proposition \ref{propositionCE}}
Following Proposition \ref{propositionAO}, we have:
\begin{equation*}
    \Sigma_{i,j}(\dot{\mathbf{CF}}) = 1
\end{equation*}
From Equation \eqref{eq:AO0}, we obtain:
\begin{equation*}
    \Sigma_{i,j}(\dot{\mathbf{CF_b}}) = \pi_0^C + \pi_1^C = 1
\end{equation*}

Hence: 
\begin{align*}
    \Sigma_{i,j}(\mathbf{\dot{E}}) =& \Sigma_{i,j}(\dot{\mathbf{CF}}-\dot{\mathbf{CF_b}}) \\
    =& \Sigma_{i,j}(\dot{\mathbf{CF}})-\Sigma_{i,j}(\mathbf{\dot{CF}_b}) \\
    =& 1-1 \\ 
    =& 0
\end{align*}
\QEDB

\noindent \textbf{Proof of Proposition \ref{propositionPiC}}
\begin{align*}
    \dot{P} &= P - P_{b} \\
        &= \Sigma_{i,j}(\dot{\mathbf{CF}}_{ij} \circ \mathbf{OB}_{ij} - \dot{\mathbf{CF}}_{ij} \circ \mathbf{TC}_{ij}) - \Sigma_{i,j}(\dot{\mathbf{CF_b}}_{,ij} \circ \mathbf{OB}_{ij} - \dot{\mathbf{CF_b}}_{,ij} \circ \mathbf{TC}_{ij})\\
        &= \Sigma_{i,j}\big((\dot{\mathbf{CF}}_{ij} \circ \mathbf{OB}_{ij} - \dot{\mathbf{CF}}_{ij} \circ \mathbf{TC}_{ij}) - (\dot{\mathbf{CF_b}}_{,ij} \circ \mathbf{OB}_{ij} - \dot{\mathbf{CF_b}}_{,ij} \circ \mathbf{TC}_{ij})\big)\\
        &= \Sigma_{i,j}\big((\dot{\mathbf{CF}}_{ij}-\dot{\mathbf{CF_b}}_{,ij}) \circ \mathbf{OB}_{ij} - (\dot{\mathbf{CF}}_{ij}- \dot{\mathbf{CF_b}}_{,ij}) \circ \mathbf{TC}_{ij}\big)\\
        &= \Sigma_{i,j}(\mathbf{\dot{E}}_{ij} \circ \mathbf{OB}_{ij} - \mathbf{\dot{E}}_{ij} \circ \mathbf{TC}_{ij})\\
        &= \Sigma_{i,j}\big(\mathbf{\dot{E}}_{ij} \circ (\mathbf{OB}_{ij}-\mathbf{TC}_{ij})\big)\\
        &= \Sigma_{i,j}(\mathbf{\dot{E}}_{ij} \circ \mathbf{\dot{CB}}_{ij})
\end{align*}
\QEDB

\end{document}